\documentclass[a4paper, 11pt]{article}

\usepackage{graphicx}

\usepackage[accsupp]{axessibility}  

\usepackage{cite}
\usepackage[T1]{fontenc}
\usepackage{authblk}
\usepackage{amsmath}
\usepackage[margin=1in]{geometry}

\usepackage{hyperref}

\usepackage{orcidlink}

\begin{document}

\title{Time-Resolved MNIST Dataset for Single-Photon Recognition}


\author[1]{Aleksi Suonsivu\orcidlink{0009-0005-8489-288X}}
\author[1]{Lauri Salmela\orcidlink{0000-0002-9836-7163}}
\author[2]{Edoardo Peretti\orcidlink{0009-0008-2076-0919}}
\author[1]{Leevi Uosukainen\orcidlink{0009-0006-0902-0771}}
\author[1]{Radu Ciprian Bilcu\orcidlink{0000-0001-7293-4353}}
\author[2]{Giacomo Boracchi\orcidlink{0000-0002-1650-3054}}


\affil[1]{\small Huawei Technologies Finland Oy, Tampere, Finland, \texttt{\{name.surname\}@huawei.com}}
\affil[2]{\small Politecnico di Milano, Milano, Italy, \texttt{\{name.surname\}@polimi.it}}

\date{}

\maketitle

\begin{abstract}
    Time-resolved single photon imaging is a promising imaging modality characterized by the unique capability of timestamping the arrivals of single photons.
    Single-Photon Avalanche Diodes (SPADs) are the leading technology for implementing modern time-resolved pixels, suitable for passive imaging with asynchronous readout.
    However, they are currently limited to small sized arrays, thus there is a lack of datasets for passive time-resolved SPAD imaging, which in turn hinders research on this peculiar imaging data.
    
    In this paper we describe a realistic simulation process for SPAD imaging, which takes into account both the stochastic nature of photon arrivals and all the noise sources involved in the acquisition process of time-resolved SPAD arrays. We have implemented this simulator in a software prototype able to generate arbitrary-sized time-resolved SPAD arrays operating in passive mode. Starting from a reference image, our simulator generates a realistic stream of timestamped photon detections.  We use our simulator to generate a time-resolved version of MNIST, which we make publicly available. Our dataset has the purpose of encouraging novel research directions in time-resolved SPAD imaging, as well as investigating the performance of CNN classifiers in extremely low-light conditions.

    \medskip
    \noindent\textbf{Keywords:} Time-Resolved Imaging $\cdot$ Simulation $\cdot$ Single-Photon Imaging $\cdot$ SPAD $\cdot$ Image Recognition $\cdot$ Low-Light
\end{abstract}

\section{Introduction}

Pixels in traditional cameras (CCD, CMOS) operate by collecting a large amount of photons during a certain period of time (the exposure time) and convert the photon flux into an electric charge, thanks to the photoelectric effect. The generated voltage, after amplification, is then passed through an analog-to-digital converter which produces the output digital raw image.
This mechanism of capturing digital images has several drawbacks: the full well capacity limits the amount of photons that can be collected during the exposure time in each pixel. The analog amplifier and digital-to-analog converter introduce noise which will affect the  quality of the output image. 
This can be observed especially in low photon flux situations where the amount of collected photons is small leading to a low signal-to-noise ratio (SNR). The way this problem is tackled, at pixel level, is by increasing the exposure time, allowing the pixel to collect more photons thus generating a stronger signal. This has an undesired effect in case of motion, since objects in the scene moving during the exposure time will be blurred into the output image. 
Moreover, in high dynamic range scenes, where some pixels are under weak photon flux while others are strongly illuminated, the captured image often contains noisy areas (due to low SNR) as well as saturated regions (due to exceeding the full well capacity of the pixel). Finally, since the photon flux is converted into an image only at the end of the exposure time, the frame rate of the sensor can't be arbitrarily increased. 

Recent advancements in electronic technologies, allowed the manufacturing of sensor units sensible to single photons, with extremely high readout speed.
Among these novel sensors, the Single-Photon Avalanche Diodes (SPADs) are able to detect the impinge of a single photon by generating an exponential gain of the voltage produced by a single photoelectron.
Time-Resolved SPADs (TR-SPADs) are SPADs supported by a fast time-to-digital converter in order to precisely timestamp the arrival of each photon, with precision in the order of picoseconds. The non-linear response of SPAD pixels enables sensing of scenes with very high dynamic range, without requiring multiple exposures or additional circuitry, indeed theoretically they saturate only with an infinite photon flux \cite{pf_spad}.
Moreover, asynchronous SPAD pixels are readout as soon as a photon is detected, without the need to wait for an exposure time to complete.
Thanks to these features, SPAD pixels are revolutionary with respect to traditional pixels, and they are attracting increasing interest from the imaging community.

In contrast with the flourishing research in models and methods for images acquired from \emph{conventional} cameras, the literature concerning time-resolved imaging is extremely limited and surprisingly scarce when it comes to deep learning techniques. On the one hand, one of the major factors inhibiting the research in time-resolved data is the lack of open-access datasets, which played a crucial role in developing models for color images \cite{deng2009imagenet} and point clouds \cite{wu20153d}. The scarcity of training sets is also due to the fact there are no high resolution asynchronous SPAD sensors in commerce (e.g., only 23 pixels in \cite{spad23}). On the other hand, time-resolved SPAD arrays pose new challenges to computer vision and learning-based algorithms, requiring new research approaches. These challenges include handling asynchronous photon arrivals, which prevents the straightforward use of algorithms meant for arrays, and the data size, which grows significantly in mid-luminance scenes acquired with non-trivial sized sensors, as raw data can easily contain thousands of timestamps per pixel. Existing reconstruction algorithms can mitigate these issues by estimating the underlying photon flux from the statistics of the photon arrivals, returning a conventional digital image. However, this approach hides raw data's asynchronous and online nature, and adds a computational step that might be unnecessary for some vision tasks.

In this work, we address the data scarcity problem by presenting a simulation procedure for generating synthetic photon detection timestamps of a time-resolved SPAD array, modeling all the relevant noise sources.
Starting from a reference image, the simulator generates a stream of photon detections whose distribution is compatible with the scene depicted by the reference image.
Additionally, the simulator can adjust the rate of photon arrivals based on a specified scene illuminance. 
We foresee two impactful use of datasets generated by our SPAD array simulator: first, enabling researching machine learning models natively processing streams of photon arrivals, second investigating established CNN performance in arbitrarily low light conditions, on the reconstructed flux.

To this end, we publicly release a dataset synthetically generated with our simulator from the traditional MNIST \cite{mnist_lenet}, at different illuminance levels.
The dataset comes in two versions: the raw time-resolved version (TR-MNIST), consisting of raw photon detections, and a collection of reconstructed images using state-of-the-art flux estimation techniques \cite{pf_spad, ip_spad} (TR-MNIST-rec). We believe TR-MNIST dataset will allow the computer vision community to design and test new models that natively process streams of photons, especially in low-light scenarios. Furthermore, TR-MNIST-rec provides a benchmark for testing the robustness of image classifiers when applied to flux estimated in extremely low-light conditions. In this regard, we have trained and tested a baseline CNN classifier on the TR-MNIST-rec dataset, assessing the classification performance drop when reducing the luminance for different flux reconstruction methods.

\section{Passive Time-Resolved Single-Photon Imaging}  

Single-Photon Avalanche Diodes (SPADs) are photodetectors able to detect the arrival of single photons to the pixel surface \cite{spad_el}.
Likewise traditional CMOS sensors, when a photon impinges a SPAD pixel, an electron is generated due to the photoelectric effect.
However, the characteristic of SPADs is that the photoelectron starts a self-sustained avalanche of charges in the semiconductor, which can be regarded as an exponential gain.
A specialized electronic component, the quenching circuit, is used to detect the avalanche and to issue a detection event. 
In case of time-resolved imaging, a time-to-digital conversion is performed as soon as the avalanche is detected, with picoseconds jitter.
SPAD sensors with time-tagging capabilities of single-photons are called \emph{Time-Resolved} SPADs (TR-SPAD).
Finally, the quenching circuit resets the sensor to a steady state, where it is ready for new detections. 
The time interval between the beginning of an avalanche and the restoration of the operating condition
is called \emph{dead time}, during which the sensor is not able to detect further photons.

Similarly to traditional CMOS cameras, SPAD detectors can be arranged in large arrays, in order to sense a wide field of view.
Arrays of TR-SPADs can operate \emph{asynchronously}, meaning that each single pixel generates an event as soon as a detection occurs, independently from the other pixels in the array. 
Small asynchronous TR-SPAD arrays are employed in fluorescence lifetime imaging microscopy \cite{async_spad_flim}, providing unprecedented imaging functionalities.
Recently, it has been shown \cite{pf_spad} that asynchronous SPAD arrays can be used to perform \emph{passive} imaging, that is to passively detect photons from the natural scene illumination, with rate limited only by the SPAD dead time.
In this work, we are interested in passive imaging with an asynchronous SPAD array, and we refer to this imaging technique as passive TR-SPAD.

Due to the asynchronous readout, different pixels will receive a different numbers of timestamps, resulting in raw data that cannot be organized in an array. Note that even uniformly illuminated pixels might yield different number of timestamps, due to the stochastic nature of photon arrivals \cite{hasinoff2021photon}. This peculiarity of passive TR-SPAD sensors calls for specialized flux estimation algorithms \cite{pf_spad, ip_spad} that allow reconstructing a digital image from the sequence of photon detections, which can be used to apply traditional image processing and computer vision algorithms to TR-SPAD data. Algorithms directly handling the photon stream for high-level vision tasks are an open research direction, which we believe will attract a lot of attention in the next years.

More formally, for each pixel $x \in X \subseteq \mathbb{Z}^2$ and given an exposure time $T \in \mathbb{R}^+$, a SPAD array yields the following raw data:
\begin{align}
&N_T(x) \in \mathbb{N} \\
&X_i(x) \in \mathbb{R}^+
\qquad
i=1,2,\dots,N_T(x)
\end{align}
where $N_T(x)$ is the number of photon-detection events at pixel $x$ during the exposure $T$, and $X_i(x)$ is the time of arrival of the $i$-th photon detected at pixel $x$.
The statistical description of these quantities is presented in Section \ref{sec:simulator}.

\subsection{Alternative paradigms for SPAD sensors}

Albeit we consider passive asynchronous TR-SPAD, there exists a few different ways to employ SPAD sensors. The simplest SPAD pixels do not have a time-to-digital converter, and can be considered as traditional pixels with full-well capacity equal to one. 
In this case, the readout is synchronous and thus the raw data consists in an array of binary pixels, also called \emph{jots}, where a jot is set to one when at least an incident photon occurs during the exposure time.
Typically, several binary images are acquired in a sequence, yielding a 3D volume of raw binary values, which are then processed to form a digital image.
These binary SPAD arrays are a possible implementation of the Quanta Image Sensor (QIS) paradigm \cite{fossum2016quanta}, which envisioned computational photography of binary volumes.
In low-light scenarios, most of the jots of a QIS sensor will be zero, while an asynchronous TR-SPAD generates a timestamp only when a detection occurs.
Thus, TR-SPAD arrays are more efficient than QIS in low-light, concerning the amount of data created.
However, at bright illumination conditions the amount of arriving photons, and thus the amount of data, can get extremely large and the data rate might have to be limited especially for large arrays of SPADs.

TR-SPAD arrays can also be readout synchronously, yielding a frame of timestamps, where each timestamp tags the first photon detection during the exposure period at the corresponding pixel.
The successful application of synchronous TR-SPAD arrays resides in \emph{active} imaging frameworks, where it is possible to relate the timestamps with the timing of an illuminating source.
For instance, time-resolved SPAD arrays and pulsed lasers are enabling technologies of recent LiDAR \cite{async_lidar} or fluorescence lifetime imaging \cite{spad_biphotonics} systems.
In these cases, a synchronous SPAD pixel detects at most one photon during each period of the illuminating laser, thus requiring several repetitions used to build an histogram of arrival times, which is then processed.
Conversely, an asynchronous TR-SPAD pixel tags a continuous stream of photons.

\section{Related Work}

In this section, we briefly review the main works related to our contribution.
We start by discussing the use of synthetic data in the TR-SPAD literature.
Then, we relate TR-SPAD to event cameras, which provide a similar stream of events. Then, we consider existing deep neural network models for stream of events, that we believe could be applied also to TR-SPAD raw data.
Finally, we discuss some limitations of existing low-light datasets that can be overcome by our simulator.

\subsection{Synthetic datasets for TR-SPAD applications}
A few synthetic TR-SPAD datasets have been previously presented for active imaging. In \cite{spad_planes, spikng_digits} a simulator is used to generate LiDAR acquisitions based on a TR-SPAD array, with the purpose of learning features extraction and classification models. Similarly, for fluorescence lifetime imaging with SPAD sensors, it is common to use simulated histograms of arrival times \cite{ann_fli} or detection timestamps \cite{rnn_fli} to train neural networks for fluorescence lifetime estimation. However, this approach is primarily concerned with the simulation of the fluorescence decay of fluorophores, and not for passive imaging of natural scenes. 
Moreover, both simulators are framed in the context of small-sized synchronous TR-SPADs arrays, with frames timed by the laser.
On the opposite, we simulate acquisitions of asynchronous photon detections for arbitrary sized SPAD arrays.

Earlier works on passive TR-SPAD \cite{pf_spad, ip_spad} made use of a TR-SPAD simulator to illustrate advantages of hypothetical TR-SPAD arrays over conventional CMOS cameras. 
These simulators receive as input the true photon flux, from which generates a stream of detection, modeling the pixel's parameters and the noise sources.
However, only a handful of simulated images were disclosed in the papers, and neither the simulator nor a dataset of simulated images have been publicly released.
We developed a simulator that receive as input any reference (grayscale or RGB) image and an arbitrary illuminance level. 
These are mapped to a photon flux, from which we simulate the stream of detections in each pixel, following the same modeling as in \cite{pf_spad}.
Moreover, we publicly release an entire dataset of simulated TR-SPAD acquisitions.

\subsection{Relation with event cameras}

Neuromorphic or Event Cameras (ECs) are imaging technologies where each pixel responds independently and asynchronously to changes in the local brightness. ECs return asynchronous events, containing the timestamp with microseconds resolution and the polarity of the brightness change. Thus, ECs' raw data is, at some extent, similar to that of asynchronous TR-SPAD arrays. However, ECs are limited to cases where the scene is  changing or the camera is moving. In contrast, a TR-SPAD camera can also be use to record still scenes, by detecting individual incoming photons, with a time resolution of picoseconds.
Recent works \cite{sodacam, generalized_ec} investigated the use of SPAD arrays for emulating ECs. 
To foster the development of learning-based solutions for ECs, several datasets were created and released to the research community.
Typical datasets for recognition were acquired in laboratory settings, for instance by displaying a moving image on a computer monitor (e.g. MNIST-DVS \cite{mnist-dvs}), or by moving the camera (e.g. MNIST-N \cite{mnist-n}).
In order to generate more realistic and diverse datasets, EC simulators have been developed.
Using the DAVIS simulator \cite{davis_simulator}, a synthetic dataset of digits, called blackboard MNIST, have been presented in \cite{acnn_detection}.
To promote the machine learning research for asynchronous passive TR-SPAD imaging, we release a Time-Resolved MNIST (TR-MNIST) dataset, by simulating the traditional MNIST \cite{mnist_lenet}, as if it had been acquired by an asynchronous passive TR-SPAD system. 
Similarly to ECs simulators, which require as input a 2D or 3D scene with a camera trajectory, our simulator receive a digital image, and simulate a stream of detected photons compatible with the input image.

\subsection{Algorithms for stream of events}

Image reconstruction algorithms are required to handle event-based data, such as TR-SPAD or EC raw data, with established CNN architectures for vision tasks.
Therefore, to avoid the reconstruction step, tailored architectures suitable for streams of events are needed. Asynchronous CNNs \cite{acnn_detection} have emerged as an interesting model for processing EC events.
In these models, the events are integrated into a feature surface, and the computation is performed only when an activation is affected by an event.
An alternative consists in the sparse CNN \cite{ascnn_ec}.

Typically, algorithms for active TR-SPAD process the histogram of photon arrivals \cite{ann_fli}, obtained by several repetitions of the synchronizing laser pulse.
Timestamp data can be processed by recurrent neural networks \cite{rnn_fli} or
Spiking Neural Networks (SNNs), which are naturally designed to handle asynchronous spikes (e.g. events).
SNNs have been investigated for fluorescence lifetime estimation \cite{snn_fli}, LiDAR \cite{ssn_lidar} and event cameras processing \cite{ssn_ec}.
In this paper, we evaluate a traditional CNN baseline with respect to different flux estimators and simulated illuminance levels.
We defer to future works the investigation of dedicated algorithms for streams of photon detections for high level tasks.

\subsection{Low-light datasets}

The other interesting aspect of our simulator is its  capability to generate images with arbitrary illuminance levels, which is particularly useful for simulating acquisitions in low-light.
In case of low-light scenarios, the leading approach in deep learning for computer vision tasks consists in first enhancing the low-light image, so that usual models can then be used.
The enhancement is in turn often performed by suitably trained deep models \cite{dl_low_survey_23}.
However, collecting real-world datasets for training models for image enhancement is a long and tedious process, which requires multiple exposures.
Indeed, some datasets do not even provide the corresponding ground truth images (e.g., ExDark \cite{exdark_dataset}).
Furthermore, datasets with paired images either contain few pairs (e.g., LOL \cite{lol_dataset}) or consist of video frames (e.g., LLIV-Phone \cite{lliv_dataset}). 

Therefore, synthetic datasets are often used to train enhancement networks. 
The typical simulation approach consists in manipulating a reference image, e.g. by gamma correction, and then introducing noise, e.g. white Gaussian noise \cite{llnet}.
These approaches are driven by simplyfing assumptions that do not necessarily match a real-world scenario.
In contrast, our simulator models the entire acquisition process of a TR-SPAD array, from photon arrivals to pixels readout, allowing to specify several sensor's parameters, thus generating more realistic acquisitions.
By using reconstruction methods, our simulator allows to generate arbitrary datasets for low-light vision tasks.

\section{Simulator}
\label{sec:simulator}

In this section, we describe the model that we use to explain the photon detection process on time-resolved SPAD (Section~\ref{sec:spad_char}), and we illustrate our simulator to generate data from a scene with different illumination levels (Section~\ref{sec:scene_to_photons}).

\subsection{SPAD Characteristics}
\label{sec:spad_char}

The photon arrivals on an imaging sensor are generally described by the Poisson statistics \cite{hasinoff2021photon}. 
The probability of detecting $N_T$ photons from a photon stream with an expected number of photons $\bar{N}_T$ (over a time interval $T$) is given by:
\begin{equation}
    P(N_T) = \frac{\bar{N}_T^{N_T}}{N_T!} \text{e}^{-\bar{N}_T}.
\end{equation}
The expected number of photons depends on the intensity of light, or the flux $\Phi$, and also on the exposure time during which the sensor collects light. The arrival times of the photons follow a random uniform distribution, conditioned on the number of arrivals \cite{hasinoff2021photon}.
This model is common in the image sensors literature \cite{timmermann1999multiscale, pf_spad, yang2011bits, wei2020physics}.
Particularly at low flux levels, or with a short exposure time, the stochastic shot-noise (resulting from the variance of the Poisson process) plays an important role when estimating the flux on the sensor.

With SPADs, the intrinsic uncertainty related to the photon statistics is accompanied by several sensor-specific factors.
Following a photon detection in a SPAD pixel, an electron avalanche is observed. This accounts for one of the main restriction in SPAD imaging, as the pixel becomes inactive for a time duration, known as {\bf dead time} $\tau_d$, during which the SPAD is quenched and recharged. The dead time in SPADs typically spans from a few to 100's of nanoseconds \cite{bronzi2015spad, gramuglia2021low}, restricting the maximum number of photons that the sensor can detect over a time interval $T$ to $\lfloor T/\tau_d \rfloor$.
 
Even though TR-SPAD sensors are capable of resolving single photon arrivals with high temporal resolution, not all of the photons arriving on the sensor are detected, even when the sensor is active. 
The {\bf quantum efficiency}, $q$, describes the detection probability for a photon that hits the active region of the SPAD detector. Note that the terms quantum efficiency and photon detection probability (PDP) are commonly used as synonyms, and in the following by quantum efficiecy we refer to the probability that a photon hitting the active area of the SPAD triggers an avalanche that is detected.
This process has a wavelength-dependence as well, with a peak detection probability in the visible wavelength range for Silicon based detectors \cite{bronzi2015spad}.

SPAD pixels suffer from spurious counts, that are detection events not corresponding to an actual incident photon.
A part of the electrons or holes may be trapped during the avalanche in a metastable state and released shortly after the initial avalanche \cite{morimoto2021scaling}. The trapped carrier may cause a secondary avalanche, an after pulse, with a probability $P_{ap}$ (typically in order of $\sim$1~\% \cite{bronzi2015spad}) and timestamp that follows an exponential decay with a mean life time $\tau = 0.5\tau_d$ \cite{wang2019afterpulsing}.
The likelihood of this \textbf{afterpulsing} effect increases when the SPAD's dead time is minimized for maximal the count rate.
Due to thermal effects and other factors, SPAD sensors exhibit another source for additional counts, even at complete darkness. These uncorrelated (Poissonian) counts are referred to as {\bf dark counts}, and their rate (DCR) typically spans from 10 to 1000's of Hz \cite{xu2017comprehensive, sicre2021dark,morimoto2021scaling, bronzi2015spad}.

Finally, regarding the timestamps recorded by a TR-SPAD detector, there is some uncertainty in the recorded photon arrival time. Due to the avalanche initiation process, the timestamps exhibit {\bf timing jitter} that can be modelled by a normal distribution with a sub-ns variation, assuming photon absorption prominently in the depletion region of the SPAD \cite{sun2019simple}.

\subsection{From scene to photons}
\label{sec:scene_to_photons}

\begin{figure}[t]
    \centering
    \includegraphics[width=0.8\textwidth]{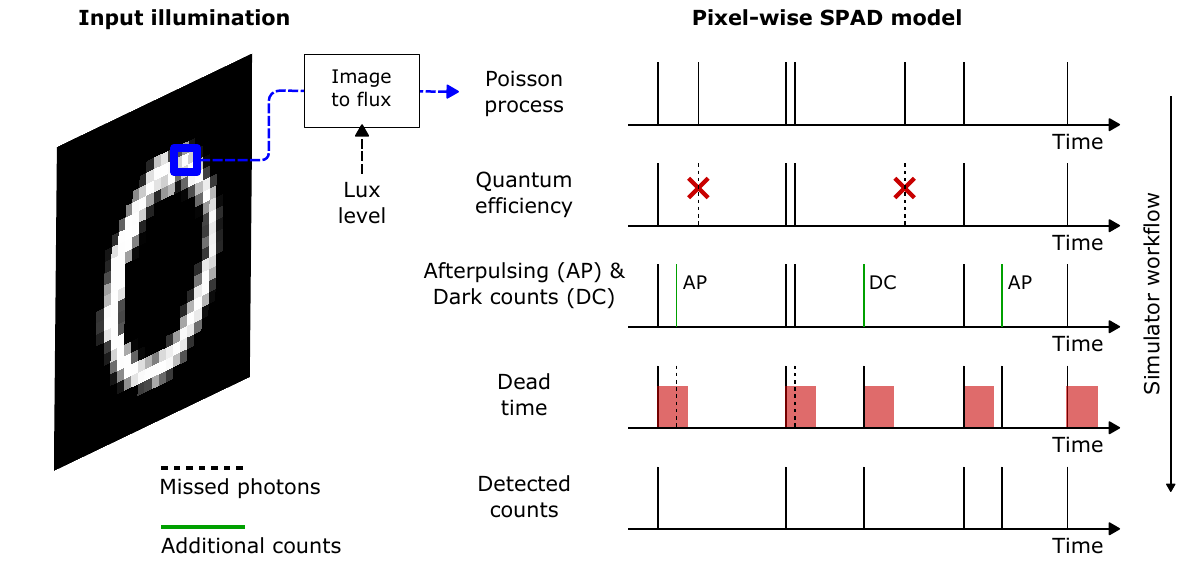}
    \caption{Illustration of pixel-wise SPAD model used by our simulator. An input image is transformed to a photon flux, via a reference lux level, which is then used to simulate photon detections. 
    The pixel-wise SPAD model is applied to all pixels in the image.}
    \label{fig:schematic}
\end{figure}

The simulation process, illustrated in Fig.~\ref{fig:schematic}, starts by defining the parameters of the SPAD and the reference scene, i.e. the intensity profile from a grayscale (or RGB) image and the reference lux level, that is the maximum lux level in the scene.
However, most images captured with standard cameras lack illumination information.
To overcome this restriction, we assign a lux value for grayscale images where the maximum brightness or "white" corresponds to a reference lux level at the sensor level. 
When considering RGB reference images, the images can be first transformed to grayscale (as no commercial TR-SPAD sensor is currently equipped with color filters) or the simulation process can be done in a per-channel manner as if the sensor had a color filter array. For grayscale images analysed in this work, we
consider a monochromatic illumination at the peak of the luminous efficacy at 555~nm, and the grayscale images are transferred from luxes to lumen values and to power on pixel (W) by
\begin{equation}
    I_W = \frac{A_p}{683 \ \text{lm/W}} I_{lux},
\end{equation}
where 683~lm/W is a transfer factor from lumens to power 
at 555~nm \cite{arecchi2007field}, $A_p$ is the pixel area and $I_{lux}$ is the input image in lux.
The expected photon flux (photons per second on pixel) can be then calculated by 
\begin{equation}
    \bar{\Phi} = \frac{I_W}{E_p},
\end{equation}
where $E_p(\lambda) = h_pc/\lambda$ is the energy of a photon with wavelength $\lambda$, and variables $h_p$ and $c$ are Planck constant and the speed of light, respectively.
This process is illustrated by the "Image to flux" block in the left part of Fig.~\ref{fig:schematic}.

The actual photon counts per pixel per exposure time $T$ are then sampled from the Poisson distribution ($N_T \sim \mathrm{Poiss}(\bar{\Phi} T)$) and the time-resolved photon data is created. Next, the quantum efficiency of the SPAD is considered by removing photons with a probability $(1-q)$ from the stream, and the dark counts are added to the stream. The after pulsing effect is then considered, and dead time of the SPAD is applied to remove all photons that arrive on the pixel within the inactive time of the SPAD pixel. 
The simulation of photon detections is visualized on the right part of Fig.~\ref{fig:schematic}.
Lastly, the timing jitter is applied on the photon arrival times.

The pixel-wise model for the SPAD introduced above is applied to each pixel of the lux-valued image, and the photon arrival times at each pixel are saved into a single file following the Address-Event-Representation (AER) \cite{zamarreno2012multicasting, peng2016bag}:
\begin{equation}
    p_{i} = {(x_{i}, y_{i}, t_{i})},
\end{equation}
where ${x_{i}}$, ${y_{i}}$ are the spatial coordinates of the \emph{i}-th photon, detected at time ${t_{i}}$.

Due to the lack of large TR-SPAD arrays, it is not possible to compare images from our simulator with real-world acquisitions. However, the equations governing our simulator have been validated in previous studies describing the TR-SPAD acquisition process, as discussed in the previous subsection.

In our simulations, we consider a SPAD pixel array with a pixel pitch of 5~\textmu m and fill factor of 100~\%, quantum efficiency $q$~=~50~\%, dead time $\tau_d$~=~50~ns, after pulsing probability $P_{ap}$~=~0.5~\%, timing jitter of 200~ps, and DCR of 100~Hz.
The SPAD is kept exposured for a time duration $T$~=~1~ms, with a varying scene illumination level.
With these simulation parameters, the amount of photon detections is kept low to study the application of CNNs for (extremely) low-light conditions and to keep the amount of data reasonable.

\section{Dataset}
\label{sec:dataset}

The stunning achievements of deep learning for computer vision would not be possible without the public availability of large datasets for training. For image classification, one of the seminal datasets has been the MNIST \cite{mnist_lenet} collection of hand-written digits. 
The original MNIST contains 28$\times$28 grayscale images, subdivided in 60000 training samples and 10000 testing samples.
Each image contains a single digit (from 0 to 9), and it is labelled with the digit it contains.
MNIST's simplicity, granted by the small image size and the few classes, sparked interest on learning-based computer vision research, even with the constrained computational resources of that time.
To give a similar contribution to the newborn computer vision field for TR-SPAD cameras, we simulate the entire MNIST, with our TR-SPAD simulator described in Section \ref{sec:simulator}.

We configure our simulator to generate a stream of asynchronous photon detection data for an arbitrary input image and for multiple simulated illumination levels, ranging from 5 to 2560 mlux.
The TR-MNIST data was generated using Matlab (Intel i9-9900 CPU, 64 GB RAM) with a run time of around 40 minutes for one lux level, including loading the MNIST images, saving the TR-SPAD data and images using three reconstruction methods. The TR-MNIST dataset consists of TR-SPAD data for 10 different lux levels, always preserving the original training and test sets.
This dataset represents a benchmark for future deep learning models able to directly process TR-SPAD streams of photon arrivals. Moreover, the same simulated streams can be used to reconstruct images by state-of-the-art flux estimator and yield an image classification benchmark for vision at very low-light levels.
We additionally release a version of TR-MNIST consisting of the reconstructed flux estimations (TR-MNIST-rec).
Both versions of the dataset can be downloaded from
\url{https://boracchi.faculty.polimi.it/Projects}
(14.4 and 1.68 GB). 

\subsection{Image classification and photons statistics at low-light}
\label{ssec:recon}
To visualize the TR-MNIST dataset and perform classification, we consider the following three flux reconstruction methods:
\begin{equation}
\widehat{\Phi}_{C} = \frac{N_T}{q T}
\hspace{5em}
\widehat{\Phi}_{PF} = \frac{N_T}{q (T - N_T \tau_d)}
\label{counts_eq}
\end{equation}

\begin{equation}
\widehat{\Phi}_{IP} = \frac{1}{q} \frac{N_T - 1}{X_{N_T} - X_1 - (N_T - 1) \tau_d}.
\label{ip_eq}
\end{equation}
Estimators $\widehat{\Phi}_{C}$ ("counts") and $\widehat{\Phi}_{PF}$ ("passive free-running") in \eqref{counts_eq} are flux estimators based on the number $N_T$ of photon detected during the integration time $T$ \cite{pf_spad}.
$\widehat{\Phi}_{C}$ refers to simple photon counting regime where the flux is estimated based on the number of photons detected on each pixel, discounted by the quantum efficiency $q$. The "passive free-running" approach, $\widehat{\Phi}_{PF}$, is based on photon counting as well but it takes into account SPAD's dead time $\tau_d$. The difference between $\widehat{\Phi}_{C}$ and $\widehat{\Phi}_{PF}$ flux estimators becomes evident at high flux levels where the dead time starts to have a significant influence on the detected number of photons by reducing the time when the SPAD is active.
The estimator $\widehat{\Phi}_{IP}$ ("inter-photon") in \eqref{ip_eq} is based on the arrival times of the photons \cite{ip_spad}, where $X_1$ and $X_{N_T}$ refer to the timestamps of the first and last detected photons.

\begin{figure}[t]
    \centering
    \includegraphics[width=0.8\textwidth]{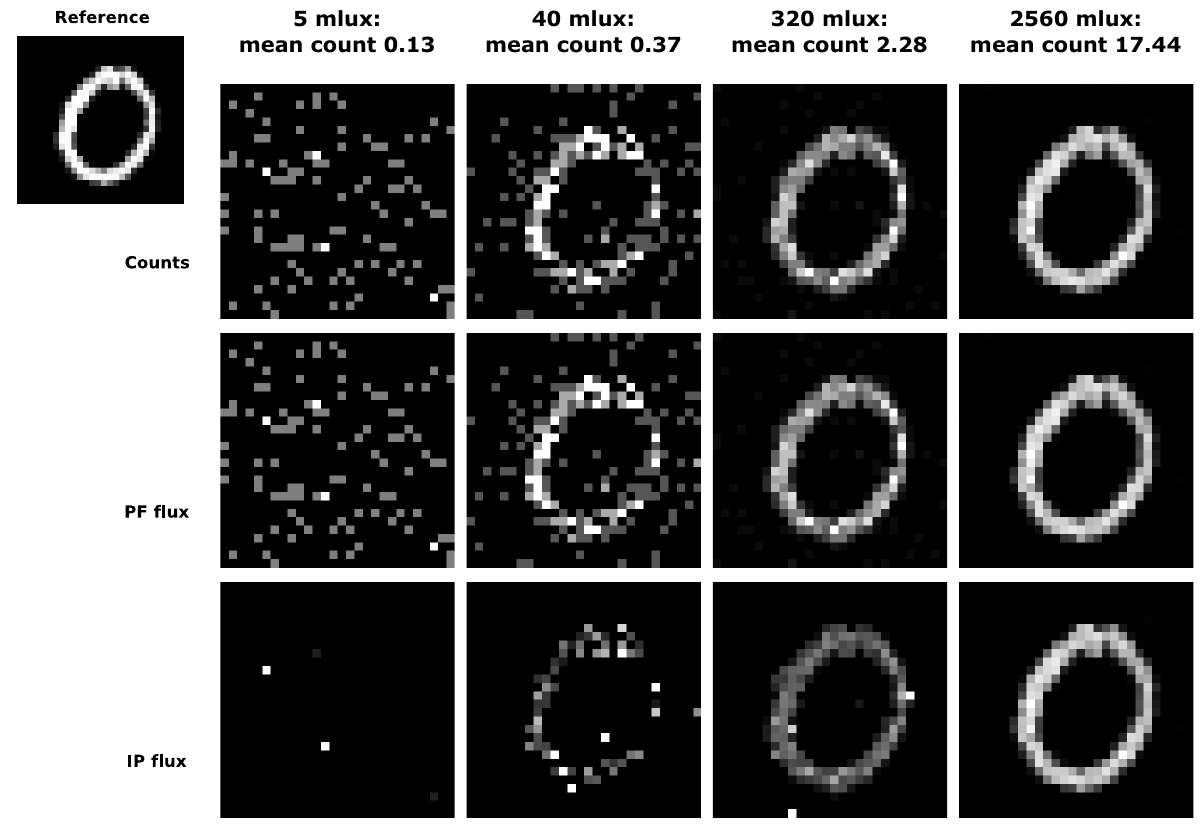} 
    \caption{Samples from TR-MNIST-rec of reconstructions at different lux levels with 1~ms integration time. All other simulation parameters have been fixed in Section \ref{sec:simulator}. 
    }
    \label{fig:examples}
\end{figure}

Fig.~\ref{fig:examples} shows examples of reconstructions based on Eq.~\eqref{counts_eq}--\eqref{ip_eq}.
We show, for four different lux levels from very low light (5 mlux) to low light (2560 mlux), three reconstructions based on the simulated photon arrivals and the original reference MNIST image (top left).
Notice that the IP flux estimator requires a minimum of two photons per pixel, and having only two counts makes the flux estimation prone to outliers. This can result into extremely high flux estimations in the dark regions of the images where essentially only dark counts are observed. Therefore, the maximum in the figures is limited to a value that is 3 times the median of the non-zero values over the training set (see normalization in the following paragraph).
The titles of each column (for one lux level) also display the mean photon count in the image.
Notice that these simulation parameters, namely DCR~=~100~Hz and $T$~=~1~ms, yield a mean dark counts of 0.1 per pixel, regardless of the lux level. Thus, for the lowest lux level of 5~mlux, the photon detections are mostly due to random dark counts.

The images in the TR-MNIST-rec dataset (original MNIST and reconstructed flux estimates) have been normalized with respect to the median value of the non-zero elements over the whole training set. This normalization scheme, in contrast to min-max normalization, is less sensitive to outliers arising from highly varying flux estimations.

\begin{figure}[t]
    \centering
    \includegraphics[width=\textwidth]{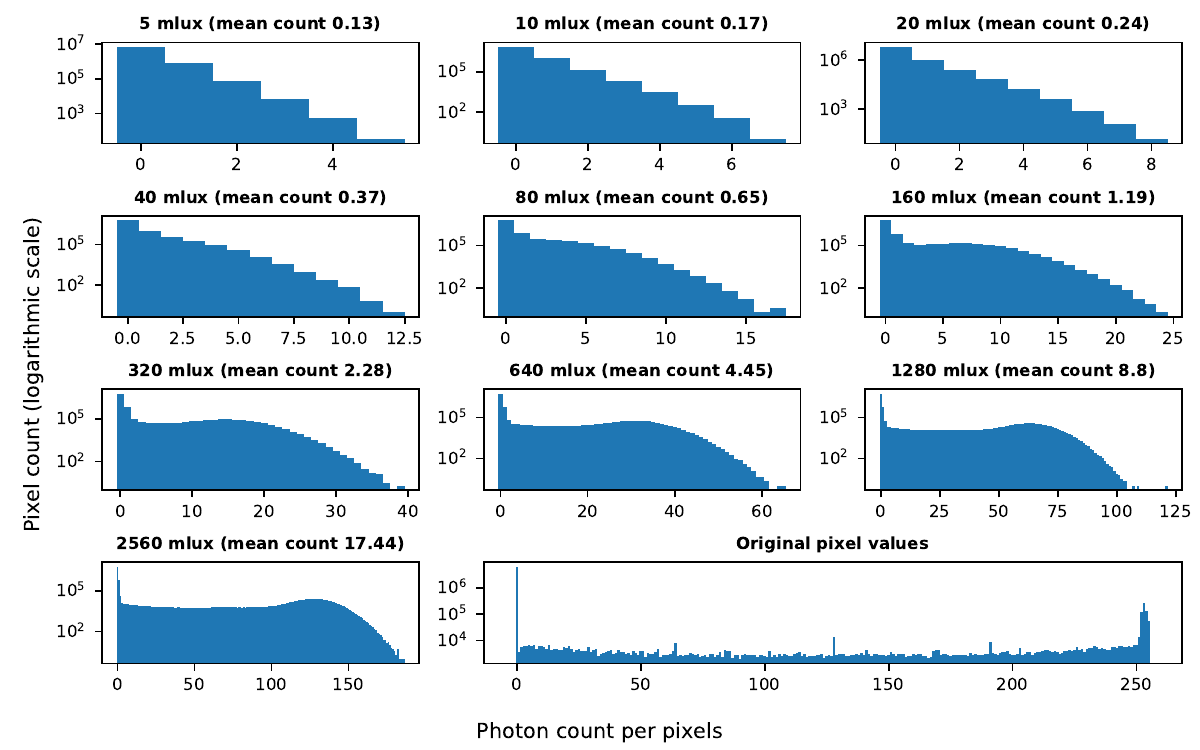}
    \caption{Histograms of photon counts at different lux levels with 1~ms integration time.}
    \label{fig:histo}
\end{figure}

Fig.~\ref{fig:histo} 
shows, for all lux levels, the distribution of pixel counts (over the 10000 test samples) in a logarithmic histogram where the horizontal axis accounts for the number of photons collected per pixel. We can see that for all the lux levels, most pixels collect no photons due to the black background of the MNIST images. The last row reports also the pixel values for the original MNIST data, with a majority of the pixels being black ("0") or white ("255"). At lower lux levels, distribution is highly concentrated at low photon counts with a tail resulting from rare multiphoton observations. However, as the lux level is increased, a secondary peak appears, originating from the white pixels of the MNIST data. The width of this peak is consistent with the Poissonian nature of the photons. In principle, the results from an extremely long exposure time would lead into statistics close to that of the original data.

\section{Baseline and Experiments}
\label{sec:exp}

The original MNIST dataset was released alongside a simple CNN, called LeNet \cite{mnist_lenet}, which was successfully trained to classify MNIST images.
LeNet is a shallow CNN consisting of only two convolutional layers having $5 \times 5$ filters and sigmoid activation functions. Two $2 \times 2$ average pooling were used to reduce spatial extent of activations. Finally, three dense layers map the convolutional features to the 10 output neurons.
 
In this section, we evaluate the same LeNet architecture as a baseline classifier on the reconstructed fluxes of the low-light TR-MNIST-rec. 
While more recent network architectures have been shown to perform better on classification tasks, we are considering LeNet as the reference CNN architecture on MNIST classification, as since it performs very well in digit recognition. This will serve as a baseline for future research on TR-SPAD classification.

We trained the LeNet for 10 epochs with batch size of 64, using Adam with learning rate of 0.001. With these training parameters, LeNet achieves accuracy of 99.22\% when trained and evaluated on the original MNIST dataset, using the same normalization scheme as for the flux estimators. The same network architecture was trained and evaluated separately for each reconstruction method and lux level pair. We chose this approach to investigate the combined impact of different photon fluxes and reconstruction methods to the classification accuracy. 

\begin{figure}[t]
    \centering
    \includegraphics[width=0.8\textwidth]{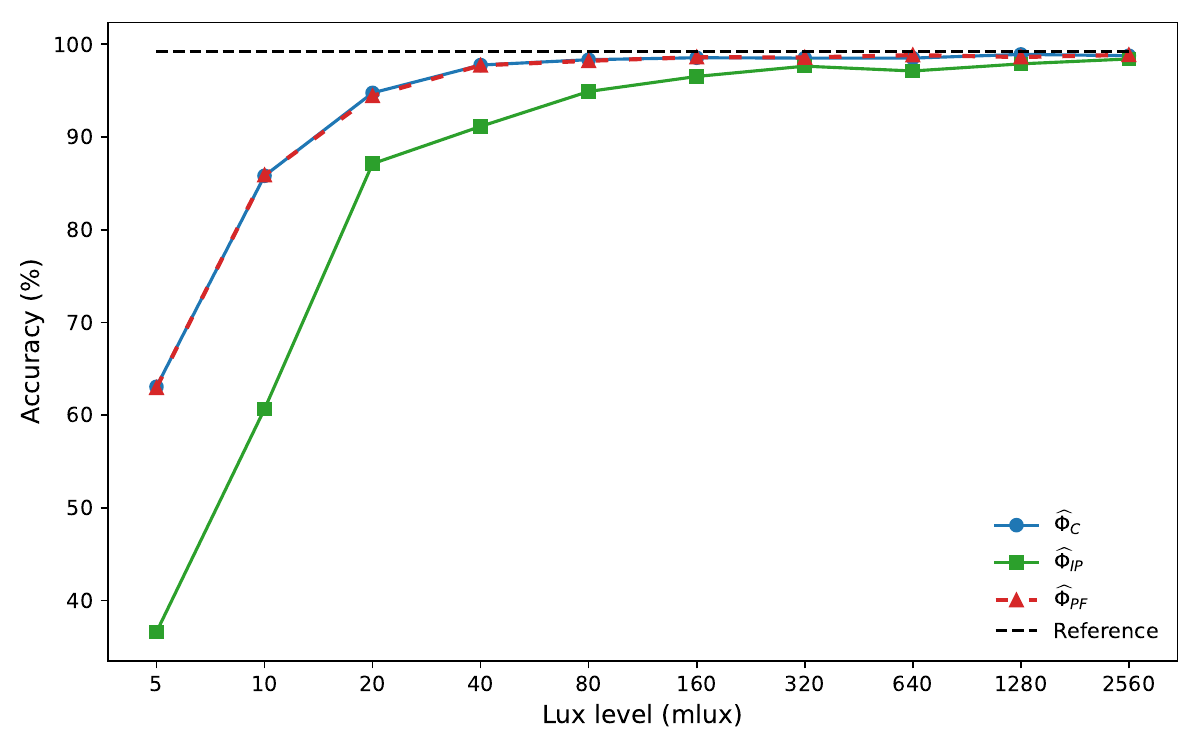}
    \caption{Classification accuracy of LeNet on TR-SPAD-rec, with different reconstruction methods and at different lux levels. The reference accuracy is obtained from the original MNIST dataset. Note that the accuracy of $\widehat{\Phi}_c$ and $\widehat{\Phi}_{PF}$ are almost coincident.}
    \label{fig:accu}
\end{figure}

Fig.~\ref{fig:accu} shows the classification accuracies with respect to the lux level and the reconstruction method. As can be seen from the plot, the performance of counts ($\widehat{\Phi}_{C}$) and PF-SPAD ($\widehat{\Phi}_{PF}$) are almost identical, while the performance of IP-SPAD ($\widehat{\Phi}_{IP}$) is the worst. Out of the three reconstruction methods, the IP-SPAD is most sensitive to noise, especially in lower lux levels. For example, if a pixel detects only two photons inside the integration time window, depending on the inter-photon arrival time, the constructed flux can vary drastically. 

As the background of the MNIST images is completely black, increasing the flux does not affect the photon counts in the black background. Therefore, the IP-SPAD reconstruction still suffers from large variance in flux estimations in the dark regions at higher flux levels as well. The reduced accuracy at high flux levels compared to other reconstructions is due to the outliers originating from the extremely high flux estimations from the background (see e.g. Fig.~\ref{fig:examples}: 320~mlux) that are present at the highest lux levels as well. Our normalization scheme eases this issue compared to min-max approach, but yet, at very low lux levels the IP-SPAD reaches lower classification accuracy. In our experiments, at around 160 mlux, all reconstruction methods reach accuracy close to the reference MNIST dataset. As the photon detection count increases, the impact of noise decreases and the accuracy improves. The classification from the simple photon counting $\hat{\Phi}_C$ and the PF-SPAD $\hat{\Phi}_{PF}$ yield similar results since the dead time of the SPAD has little effect on the estimated photon flux at these illumination levels.
Thus, the reconstructed fluxes are almost equivalent in these two cases.

\section{Conclusion}

We presented a realistic simulator for asynchronous TR-SPAD imaging sensors, and provided a benchmark for assessing image classification of TR-SPAD data.
The simulator models all the relevant characteristics and parameters of an asynchronous TR-SPAD array, with particular emphasis on all the sources of noise affecting the acquisitions. 
Its output is a stream of photon detections that can be explained by a given reference image.
Moreover, in our simulator it is possible to set the ambient lux level, in order to generate several acquisitions of the same scene under different luminance, exploring also very low-light conditions. 

We demonstrate the effectiveness of our simulator by generating a time-resolved and single-photon version of the seminal MNIST dataset.
By simulating the dataset's images at different lux levels, we provide a means to investigate the classification performance at very low-light, where the signal is heavily corrupted by noise.
We publicly release two version of this dataset to the community: the raw photon detection timestamps (TR-MNIST) and the images reconstructed with three different methods (TR-MNIST-rec).

We hope the release of TR-MNIST dataset will spark interest in the research about suitable models that can directly handle the stream of photons, avoiding the reconstruction step.
Additionally, we believe TR-MNIST-rec could be of interest for investigating classification algorithms in the challenging low-light scenarios, which is characterized by low SNR.

Our first experiments, based on the LeNet architecture, serve as a baseline for future ad-hoc photon-streams classifier and as a starting point for a systematic investigation of low-light classification.
Furthermore, we found that flux estimators based on photon counts are more robust in very low-light compared to arrival time-based estimators.

As future work, we plan to investigate ad--hoc models for image classification from the corresponding photon streams of the TR-MNIST dataset. We also plan to simulate and release more representative datasets for visual recognition problems, considering also natural RGB images.

\newpage


%
%
\bibliographystyle{splncs04}
\bibliography{main}
\end{document}